\documentclass[11pt,a4paper]{article}

\usepackage[T1]{fontenc}
\usepackage[utf8]{inputenc}
\usepackage{lmodern}
\usepackage{microtype}
\usepackage[margin=2.5cm]{geometry}

\usepackage[numbers,sort&compress]{natbib}   % Vancouver-style numbered citations
\usepackage{graphicx}
\usepackage{booktabs}
\usepackage{multirow}
\usepackage{array}
\usepackage{amsmath}
\usepackage{amssymb}
\usepackage{placeins}                       % \FloatBarrier
\usepackage[hyphens]{url}
\usepackage[hidelinks]{hyperref}
\usepackage{orcidlink}
\usepackage{authblk}

\graphicspath{{./}{./figures/}}

\title{Extraction of clinical findings from mammography and breast
  ultrasound reports: a comparison between specialists and Artificial Intelligence}

\author[1,$*$]{Lorenzo Farias\,\orcidlink{0009-0000-2219-1889}}
\author[1]{Hanna Reckziegel\,\orcidlink{0009-0001-8165-131X}}
\author[1,2]{Daniela Duarte da Silva Bagatini\,\orcidlink{0000-0001-8548-5841}}
\author[1]{Daniel Schulz\,\orcidlink{0009-0006-7266-0552}}
\author[3]{Gabriela de Andrade Monteiro\,\orcidlink{0009-0000-8294-9022}}
\author[4]{Letícia Zanatta\,\orcidlink{0009-0009-4911-3894}}
\author[5]{Ana Laura Brill Thum\,\orcidlink{0009-0004-2734-9322}}
\author[5]{Priscila Schmidt Lora\,\orcidlink{0000-0002-1543-769X}}
\author[3]{Débora Oliveira da Silva\,\orcidlink{0000-0001-7023-4927}}
\author[6]{Ana Paula Wernz da Cunha Müller\,\orcidlink{0000-0002-1919-7152}}
\author[7]{Cristiane Drebes Pedron\,\orcidlink{0000-0002-9920-3830}}

\affil[1]{Department of Engineering, Architecture and Computing, Universidade de Santa Cruz do Sul (UNISC), Santa Cruz do Sul, RS, Brazil}
\affil[2]{Graduate Program in Industrial Systems and Processes (PPGSPI), Universidade de Santa Cruz do Sul (UNISC), Santa Cruz do Sul, RS, Brazil}
\affil[3]{Graduate Program in Production and Systems Engineering (PPGEPS), Universidade do Vale do Rio dos Sinos (UNISINOS), São Leopoldo, RS, Brazil}
\affil[4]{Faculty of Medicine (FaMed), Universidade Federal do Rio Grande (FURG), Rio Grande, RS, Brazil}
\affil[5]{School of Health, Universidade do Vale do Rio dos Sinos (UNISINOS), São Leopoldo, RS, Brazil}
\affil[6]{Clínica de Mastologia Dra.\ Ana Paula Muller Ltda., Porto Alegre, RS, Brazil}
\affil[7]{Graduate Program in Administration (PPGA) and Graduate Program in Project Management (PPGP), Universidade Nove de Julho (UNINOVE), São Paulo, SP, Brazil}
\affil[$*$]{Corresponding author: \href{mailto:fariaslorenzo@mx2.unisc.br}{fariaslorenzo@mx2.unisc.br}}

\date{}

\begin{document}
\maketitle

\begin{abstract}
Breast cancer is the leading cause of cancer-related death among women in
Brazil, and the time between the request and the release of mammography reports
directly influences adherence to screening, making the agility in processing
these reports a critical factor for early diagnosis. In this context, this study
compares the performance of a Large Language Model (LLM) with manual extraction
performed by a team of health researchers in identifying clinical findings from
mammography and breast ultrasound reports written in Brazilian Portuguese.
Named Entity Recognition (NER) was applied through Prompt Engineering using a
\textit{few-shot} strategy, employing the Gemini~2.5~Flash model, selected from
preliminary exploratory tests with four candidate models. The Gemini~2.5~Flash
model demonstrated the best performance, achieving a Macro~F1 of~0.91 and a
Micro~F1 of~0.98. The subjective validation, in which 29 exams --- of different
formats --- were evaluated by health researchers using a Likert scale, yielded
an agreement index of~93.1\%. The model outperformed human extraction in
overall Macro~F1 (0.91 vs.\ 0.72), as in four reports the model correctly
identified information that had been omitted or incorrectly recorded during
manual extraction, demonstrating its potential as a complementary verification
tool alongside specialists. The results confirm the hypothesis that LLMs, when
instructed through Prompt Engineering, can achieve performance comparable to or
superior to manual extraction by health professionals.
\end{abstract}


%% Highlights (Elsevier-only, kept for reference)
% \begin{highlights}
% \item Few-shot LLMs reach 0.91 Macro F1 in clinical data extraction from mammography.
% \item AI outperformed human researches in Macro F1 by identifying missed clinical cases.
% \item Using plain text (.txt) reduces AI hallucinations compared to complex PDF layouts.
% \end{highlights}

%% ============================================================
\section{Introduction}\label{sec:intro}
%% ============================================================

Breast cancer is the leading cause of cancer-related death among women in
Brazil, accounting for 16.5\% of oncological deaths in the country~\cite{inca2024}.
The estimate for the 2023--2025 triennium is 73,610 new cases annually, with
an adjusted incidence rate of 41.89 cases per 100,000 women~\cite{inca2022}. In
Rio Grande do Sul, 5,038 cases were recorded in 2023, a 35.4\% increase over
the estimated value, representing the third highest number in the
country~\cite{observatoriocancer2025, vasconcellos2024}.

In the breast cancer care pathway, the time between the request and the release
of reports directly influences adherence to screening. In 2023, only 48.8\% of
screening mammography reports were released within 30~days, while approximately
36\% took more than 60~days~\cite{inca2024}. Furthermore, the lack of
interoperability between health systems compromises continuity of care and may
negatively affect clinical outcomes~\cite{tenoriofilho2024, costa2025,
brasil2026}. This scenario highlights the need for process improvements to
expedite feedback to patients, thereby strengthening confidence in screening
programs.

The use of Artificial Intelligence (AI) can bring numerous benefits, such as
resource optimization, faster diagnoses, and error reduction~\cite{braga2024,
leite2024, rahman2024, oliveira2024}. One approach to mitigating the latency
between examination and patient comprehension of results consists of automating
the extraction of relevant clinical findings from diagnostic reports, using the
concept of Named Entity Recognition (NER). In the healthcare context, NER
automatically extracts information such as diagnoses, medications, or dates from
reports and medical records, supporting data structuring.

Considering this context, the present study is based on the hypothesis that
LLMs, when correctly instructed through Prompt Engineering techniques, can
achieve performance comparable to manual extraction of clinical findings
performed by specialists. The selection of the LLM model used was conducted
through preliminary exploratory tests, detailed in the Methods section.

The main objective of this work is to compare the performance of an LLM model
with manual extraction performed by health researchers in identifying
pre-selected clinical findings in mammography and breast ultrasound reports.
To this end, the following specific objectives were defined: (1)~mapping,
together with health researchers and a specialist, the relevant clinical
finding present in the reports; (2)~developing and refining a structured
\textit{prompt} based on identified patterns; (3)~quantitatively evaluating
the automated extraction through classification metrics (Accuracy, Precision,
Recall, and F1-Score); and (4)~qualitatively validating the results with health
professionals through the Likert scale.

This article contributes to the field of digital health by providing empirical
evidence that LLMs, when adequately instructed, can serve as a clinical support
tool for structuring data from non-standardized reports written in Brazilian
Portuguese.

%% ============================================================
\section{Related Work}\label{sec:relacionados}
%% ============================================================

AI-based solutions have been proposed for the automated extraction of clinical
data, notably BioBERT~\cite{lee2020}, which employs the BERT (\textit{Bidirectional
Encoder Representations from Transformers}) architecture, and other NLP
techniques. However, these approaches require extensive training datasets with
local clinical terminology, which constitutes a considerable challenge for
different linguistic and cultural contexts.

Currently, LLMs demonstrate excellent results across multiple languages and
greater adaptability with less need for specific training~\cite{akcali2025,
godoy2024, mou2024, liu2025, chen2024}. Godoy~\textit{et al.}~\cite{godoy2024} evaluated three
OpenAI models (GPT-4o, GPT-4-turbo, and GPT-3.5-turbo) for automatic
laterality detection in 377~mammography reports in Spanish, using
\textit{zero-shot}, \textit{one-shot}, and \textit{few-shot} strategies.
GPT-4o with the \textit{few-shot} approach achieved an F1 of~0.77,
significantly outperforming GPT-3.5-turbo. The analysis identified three types
of errors: positioning (Type~1), format (Type~2), and severe non-correctable
errors (Type~3).

Chen~\textit{et al.}~\cite{chen2024} investigated three Chinese LLMs with
fine-tuning via LoRA for structuring 550~breast ultrasound reports. Baichuan2-7b
achieved 97.63\% precision for structure extraction and 96.22\% for
attribute-value matching. The study demonstrated that dividing reports into
logical sections significantly improved performance compared to holistic
structuring, with 400~training samples being the optimal volume.

Akcali~\textit{et al.}~\cite{akcali2025} applied a many-shot prompt engineering
strategy using Google's Gemini~1.5~Pro to automatically extract five clinical entities
--- anatomy, impression, and observation presence, absence, and uncertainty ---
from 85~Turkish mammography reports, a language with limited available NLP
resources. The method incorporated 165~annotated examples within a
26,000-token prompt, yielding a macro-averaged F1 of~0.99 under relaxed
matching and~0.84 under exact matching. The study demonstrated that
many-shot prompting with a large-context-window LLM consistently outperforms
zero-shot and few-shot strategies for non-English medical report annotation,
reinforcing the adaptability of prompt engineering to languages beyond English
without requiring task-specific model training.

%% ============================================================
\section{Methods and Procedures}\label{sec:metodos}
%% ============================================================

This study adopts an iterative experimental design, combining quantitative
performance evaluation with qualitative expert validation. Clinical variables
were defined in collaboration with health researchers and a mastologist. The
\textit{prompt} development followed an incremental process --- evaluating
multiple LLM models, input formats, and prompting strategies --- to identify
the best-performing configuration. The final system was evaluated using
classification metrics (Accuracy, Precision, Recall, and F1-Score) against a
manually adjudicated ground truth, complemented by qualitative validation via
Likert scale. The project follows the \textit{Design Science Research} (DSR)
approach~\cite{hevner2004, peffers2007}, structured in five
stages~\cite{vaishnavi2021}, described in the following subsections.

\subsection{Procedures}\label{sec:procedimentos}

\textbf{Stage~1 -- Problem Awareness:} a literature review was conducted in
PubMed, Scopus, Web of Science, and SciELO databases, which revealed gaps in
the automated extraction of clinical findings from mammography reports in
Brazilian Portuguese and the relevance of report release delays for screening
adherence. Based on this review, the breast cancer care pathway at the
screening stage was selected as the focus of this study.

\textbf{Stage~2 -- Suggestion:} mapping, together with health professionals
(mastologist and medical students), of clinical variables (cysts, nodules,
calcifications, microcalcifications, and BI-RADS) and AI technologies (RAG,
Prompt Engineering, Fine-Tuning, NLP, and LLMs).

\textbf{Stage~3 -- Development:} exploratory tests evaluating (1)~PDF vs.\
TXT formats~\cite{ji2023}; (2)~\textit{Zero-shot}, \textit{Few-shot}, and
\textit{Multi-shot} strategies; and (3)~GPT-4o, Claude~3.5~Sonnet,
DeepSeek~R1, and Gemini~2.5~Flash models. RAG was compared, but Prompt
Engineering demonstrated sufficient efficacy with lower complexity. A
Python~3.13 application was developed for extraction in structured JSON.

\textbf{Stage~4 -- Evaluation:} the evaluation was conducted by comparing
the results obtained by the AI with those produced by the researchers, using
quantitative metrics (Accuracy, Sensitivity, Specificity, and F1-Score) and a
qualitative measure represented by the Likert Scale. The F1-Score was calculated
for both human evaluators and the AI, while the Likert Scale was used to measure
the degree of agreement of the health professional with the extractions
performed by the model.

\textbf{Stage~5 -- Conclusion and Communication:} results reported in this
article.

\subsection{Study Sample}\label{sec:amostra}

The sample consisted of 33 examinations: 29 for testing and comparing
\textit{prompts}, and 4 selected as representative examples to compose the
final \textit{prompt} in the chosen \textit{few-shot} approach. The
examinations were obtained from the mastology clinic partnering with the
project, covering reports from different laboratories and radiology services
in Rio Grande do Sul, with variations in terminology and writing style, which increases the diversity of the sample.

\subsection{Data Mapping}\label{sec:desenvolvimento}

\begin{figure}[htbp]
  \centering
  \includegraphics[width=\linewidth]{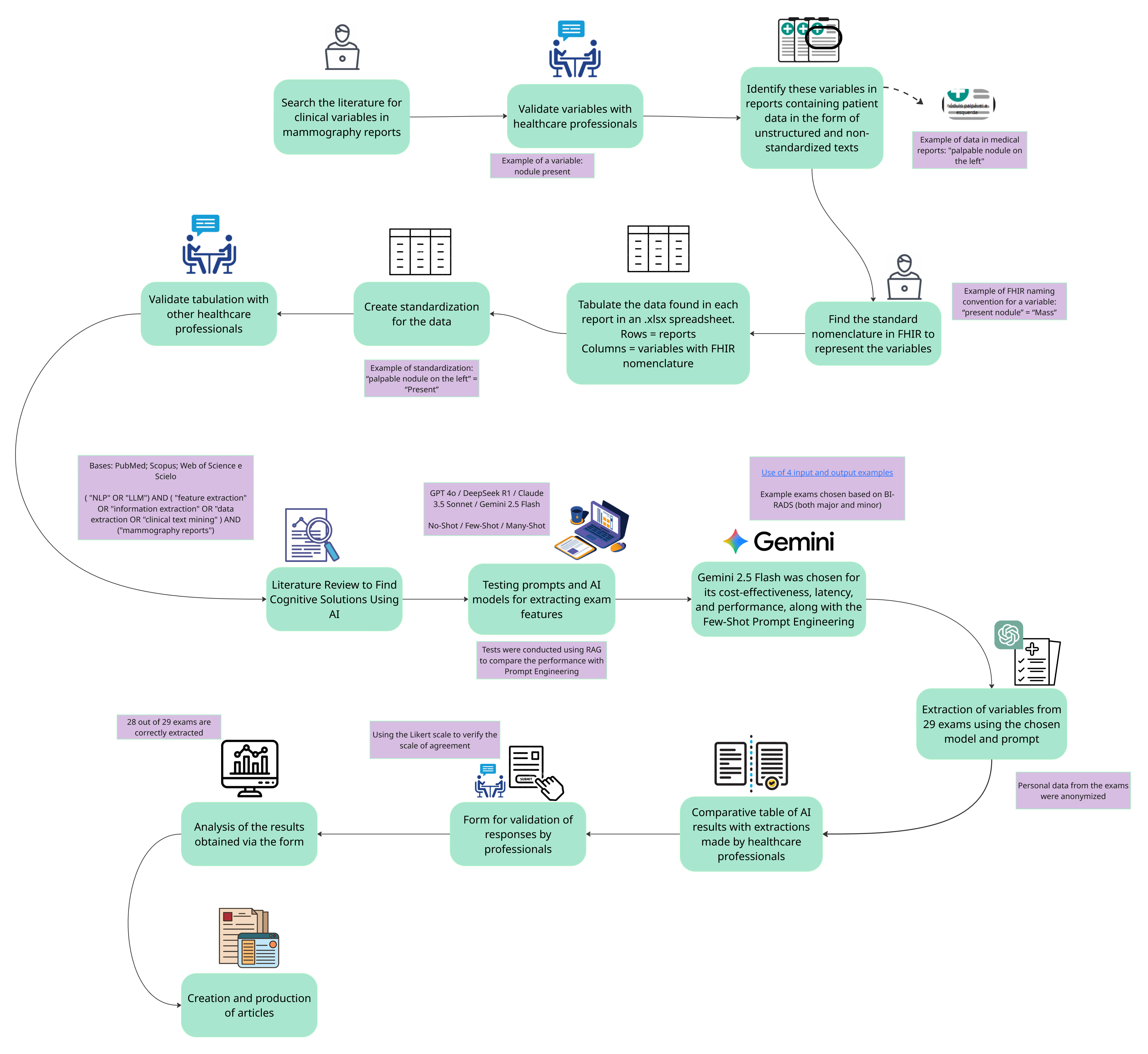}
  \caption{Mapping process flowchart. Illustrates the stages from
    bibliographic search for clinical variables, validation with health
    professionals, identification in reports, FHIR standardization, tabulation
    in spreadsheet, and final validation.}
  \label{fig:fig1}
\end{figure}

Figure~\ref{fig:fig1} illustrates the clinical data mapping stage from
examination reports. The collected reports were subjected to analysis by health
researchers, who performed the extraction of relevant findings to construct a
structured reference informational model in a spreadsheet. This process
underwent agreement evaluation between the specialist and health researchers to
ensure consistency and reliability of the extracted data.

\FloatBarrier

\subsubsection{AI Model Selection}\label{sec:selecao_modelo}

Model selection was conducted through exploratory tests with four candidates
evaluated for qualitative performance, API access cost, response latency, and
score on the LMArena benchmark~\cite{lmarena2025}, a platform that compares
models for general language processing capability. Table~\ref{tab:modelos}
summarizes the characteristics of the evaluated models.

\begin{table}[htbp]
\centering
\caption{Characteristics of models evaluated in exploratory tests.
         Prices in USD per million tokens.
         \textit{Benchmark}: LMArena score~\cite{lmarena2025}.}
\label{tab:modelos}
\begin{tabular}{@{} lcccc @{}}
\toprule
\textbf{Model}
  & \textbf{Benchmark}
  & \textbf{Latency (s)}
  & \textbf{Input P.}
  & \textbf{Output P.} \\
\midrule
GPT-4o (OpenAI)               & 1408 & 0.29 & \$2.50 & \$10.00 \\
Claude~3.5~Sonnet (Anthropic) & 1283 & 0.95 & \$3.00 & \$15.00 \\
DeepSeek~R1                   & 1359 & 3.40 & \$1.35 & \$5.40  \\
\textbf{Gemini~2.5~Flash (Google)} & \textbf{1394} & 8.15
                              & \textbf{\$0.15} & \textbf{\$3.50} \\
\bottomrule
\end{tabular}
\end{table}

\FloatBarrier

Gemini~2.5~Flash was selected for presenting the second highest benchmark score
(1,394~points), significantly lower input cost (\$0.15/M~tokens), API access
availability through Google during the research period, and best qualitative
performance in exploratory tests. Although its average latency (8.15~s) is
higher than the other models, this factor does not constitute a relevant
restriction in the context of batch processing of clinical reports.

\subsubsection{Parameter Definition and \textit{Prompt} Testing}
\label{sec:prompts}

Three prompt engineering approaches were evaluated: (1)~\textbf{Zero-shot} --
only the task instruction, with no prior examples; (2)~\textbf{Few-shot} --
from 1 to~5 examples of analyzed reports; and (3)~\textbf{Multi-shot} --
15~representative examples. The extractions generated by the AI were tabulated
and compared to the gold standard established by the multidisciplinary team of
health researchers. After performance comparison, the \textit{few-shot} strategy with 4~examples
was selected. As observed in lightweight or distilled models (such as Gemini~2.5~Flash
or smaller local models), while \textit{zero-shot} suffered from structural inconsistency
in the JSON output, \textit{multi-shot} led to attention dilution \cite{tang2025fewshot}. 
Although these models possess large context windows, providing an excessive number of 
examples exacerbates the ``lost in the middle'' phenomenon \cite{liu2023lost} and 
overwhelms their attention mechanisms. This ``over-prompting'' degrades extraction quality and increases the hallucination 
rate \cite{tang2025fewshot}, as the model's attention is spread too thin across 
the examples, causing it to confuse or merge clinical details from the examples 
with the actual report being analyzed. Therefore, \textit{few-shot} 
presented the best technical balance.

Regarding data input, PDF files can induce hallucinations~\cite{ji2023} in AI
models due to their complex structure (metadata, tables, and images that may
be interpreted inconsistently). To mitigate this issue, the TXT format was
adopted for report input, preserving only the essential textual content.

\begin{figure}[htbp]
  \centering
  \includegraphics[width=.85\linewidth]{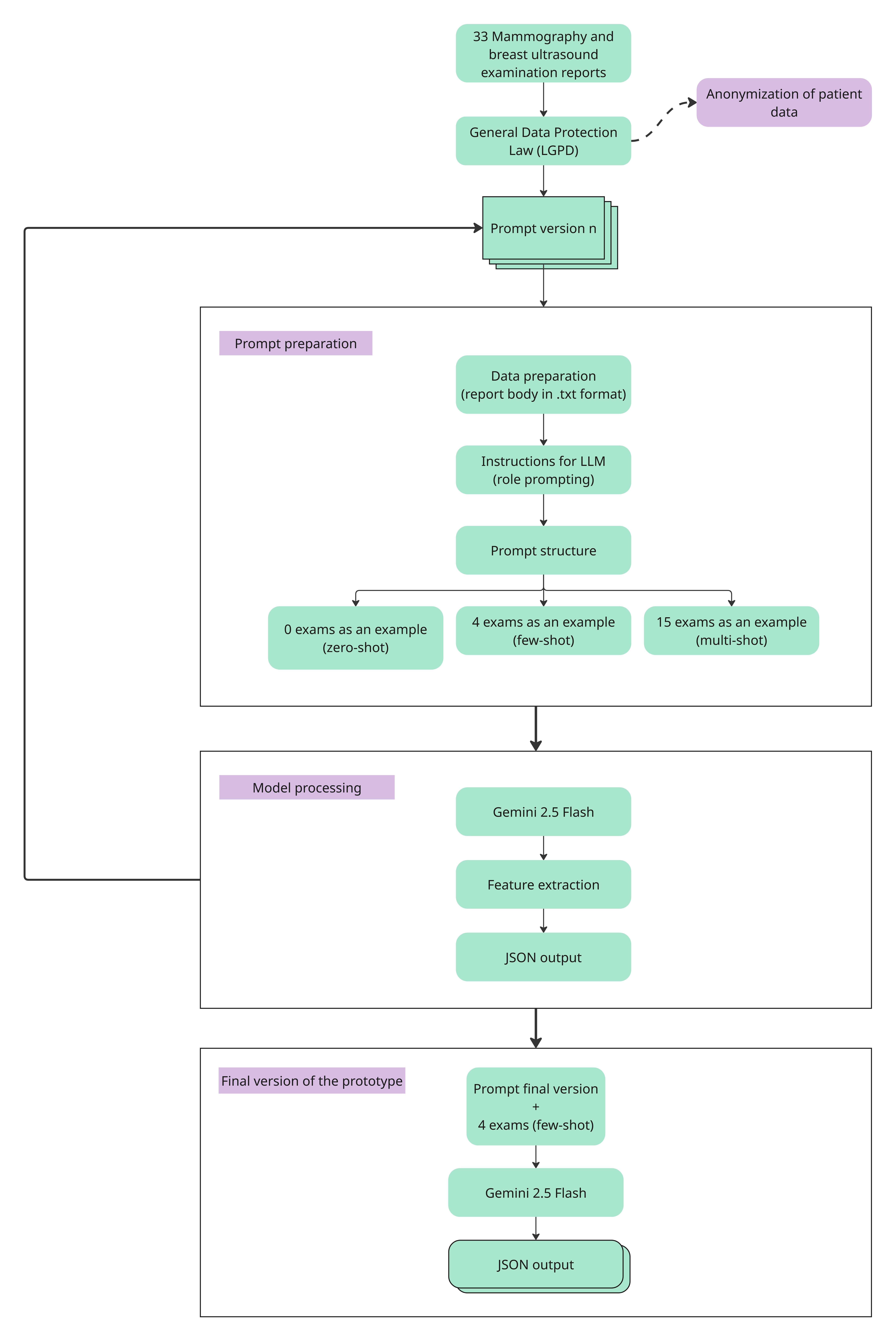}
  \caption{Technical methodology flowchart. Shows the stages from
    report anonymization to \textit{prompt} preparation, model processing, JSON
    output, and use in the final prototype.}
  \label{fig:fig2}
\end{figure}

Figure~\ref{fig:fig2} illustrates the \textit{prompt} testing workflow up to
the use of the final prototype version.

\FloatBarrier

\subsubsection{Comparative Analysis and Statistical Evaluation}
\label{sec:estatistica}

For qualitative analysis, the extractions performed by the AI were organized in
a spreadsheet, where each row represents an examination and the columns
represent clinical variables. A specialist and a health researcher received the
original reports along with the extraction spreadsheet for cross-validation.
The Likert scale was used in response to the guiding question: \textit{``Regarding
this report, are the presented answers correct?''}

For quantitative analysis, the F1-Score (harmonic mean of Precision and Recall)
was adopted as the central metric, widely used in NLP tasks~\cite{lee2020,
akcali2025}. The metrics were calculated from: \textbf{TP} (AI correctly
identifies a present finding); \textbf{TN} (AI correctly identifies an
absence); \textbf{FP} (AI identifies something nonexistent -- hallucination);
and \textbf{FN} (AI omits a present finding):

\begin{equation}
  \text{Accuracy} = \frac{TP + TN}{TP + TN + FP + FN}
  \label{eq:acuracia}
\end{equation}
\begin{equation}
  \text{Precision} = \frac{TP}{TP + FP} \qquad
  \text{Recall} = \frac{TP}{TP + FN}
  \label{eq:prec_rev}
\end{equation}
\begin{equation}
  F_1 = \frac{2 \cdot \text{Precision} \cdot \text{Recall}}
             {\text{Precision} + \text{Recall}}
  \label{eq:f1}
\end{equation}

For establishing the Ground Truth, the manual extraction by health researchers
was used, subjected to an adjudication process in cases of disagreement between
manual extraction and AI extraction. In such cases, the multidisciplinary team,
composed of the same health researchers involved in the initial extraction,
returned to the original report for joint review, determining by consensus which
result was compatible with the explicit content of the document. The ground
truth resulting from this process was adopted as the reference for calculating
all quantitative metrics.

%% ============================================================
\section{Results}\label{sec:resultados}
%% ============================================================

The performance evaluation of the Gemini~2.5~Flash model was conducted on two
fronts: objective quantitative analysis (F1-Score and confusion matrices) and
subjective qualitative analysis (agreement by researchers and specialist).

\subsection{Quantitative Performance and Adjudication}\label{sec:quant}

In the automated evaluation of 29~test reports, the model demonstrated high
efficacy in the identification and extraction of clinical entities. During
validation, it was observed that the initial Gold Standard (manual extraction)
contained inaccuracies in four examinations (IDs~1, 10, 16, and~70). After
adjudication, the AI results proved to be correct.

The system evaluated 116~binary instances (presence/absence of Cysts, Nodules,
Calcifications, and Microcalcifications in 29~reports), recording 39~True
Positives (TP), 75~True Negatives (TN), 1~False Positive (FP), and 1~False
Negative (FN). Table~\ref{tab:desempenho} presents performance by clinical
entity, with Macro~F1 of~0.91 and Micro~F1 of~0.98.

\begin{table}[htbp]
\centering
\caption{Performance per clinical entity of the Gemini~2.5~Flash
         model after gold standard adjudication.}
\label{tab:desempenho}
\begin{tabular}{@{} lccccc @{}}
\toprule
\textbf{Clinical Entity}
  & \textbf{Accuracy}
  & \textbf{Sensitiv.}
  & \textbf{Specific.}
  & \textbf{Precision}
  & \textbf{F1} \\
\midrule
Cysts                & 1.00 & 1.00 & 1.00 & 1.00 & 1.00 \\
Nodules              & 1.00 & 1.00 & 1.00 & 1.00 & 1.00 \\
Calcifications       & 0.97 & 0.94 & 1.00 & 1.00 & 0.97 \\
Microcalcifications  & 0.97 & 1.00 & 0.96 & 0.50 & 0.66 \\
\midrule
\textbf{Macro F1}    & \textbf{0.98} & \textbf{0.98}
                     & \textbf{0.99} & \textbf{0.87} & \textbf{0.91} \\
\textbf{Micro F1}\textsuperscript{a}
                     & -- & \textbf{0.98} & -- & \textbf{0.98} & \textbf{0.98} \\
\bottomrule
\end{tabular}
\par\smallskip
\footnotesize\textsuperscript{a}Micro~F1: TP\,=\,39, FP\,=\,1, FN\,=\,1
(Precision\,=\,Recall\,=\,0.975).
\end{table}

\FloatBarrier

\begin{figure}[htbp]
  \centering
  \includegraphics[width=.66\linewidth]{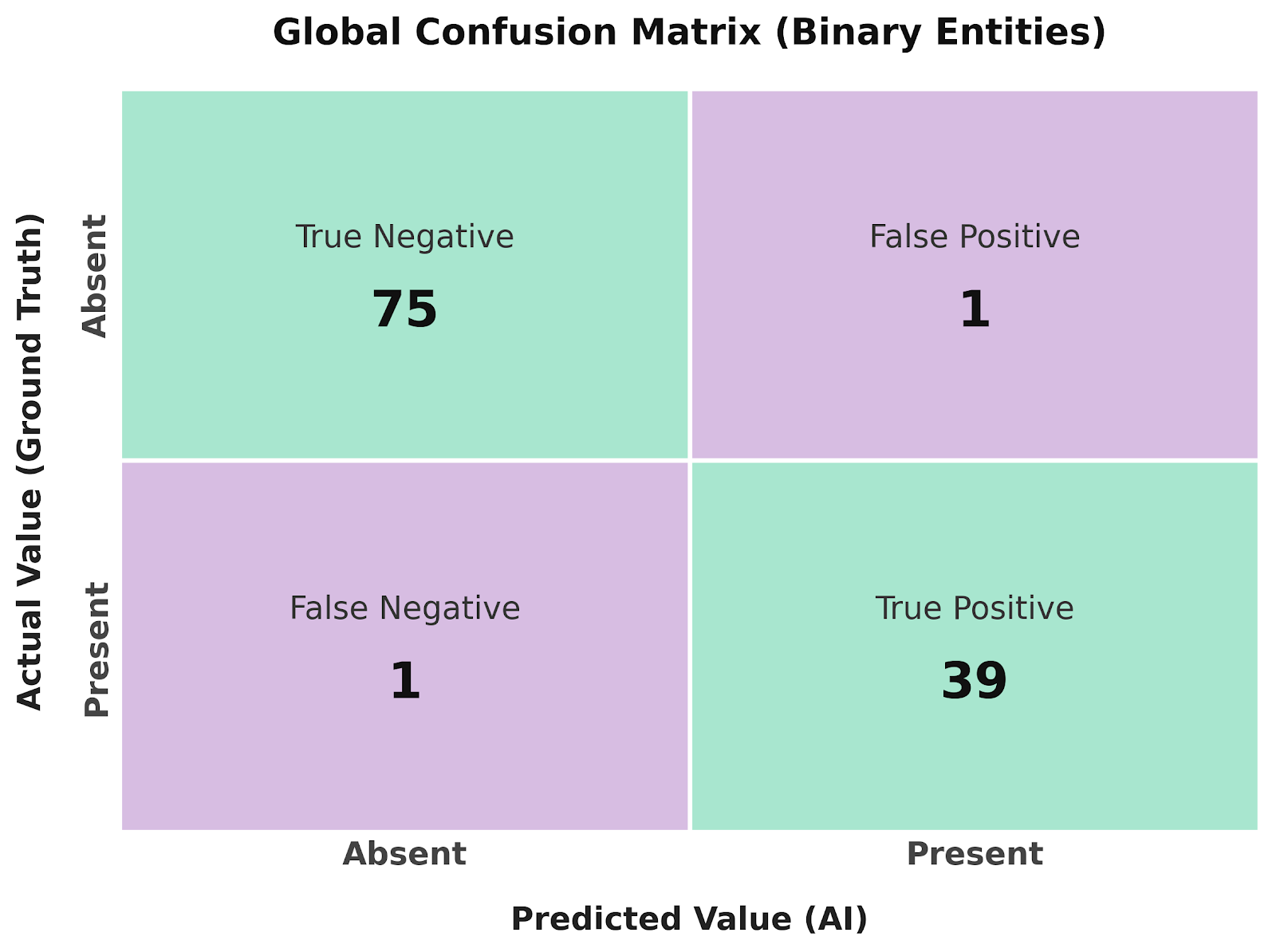}
  \caption{Global Confusion Matrix (Binary Entities).
    The model recorded 75~TN and 39~TP, with only 1~FP and 1~FN across
    116~evaluated instances.}
  \label{fig:confusao_global}
\end{figure}

\begin{figure}[htbp]
  \centering
  \includegraphics[width=.66\linewidth]{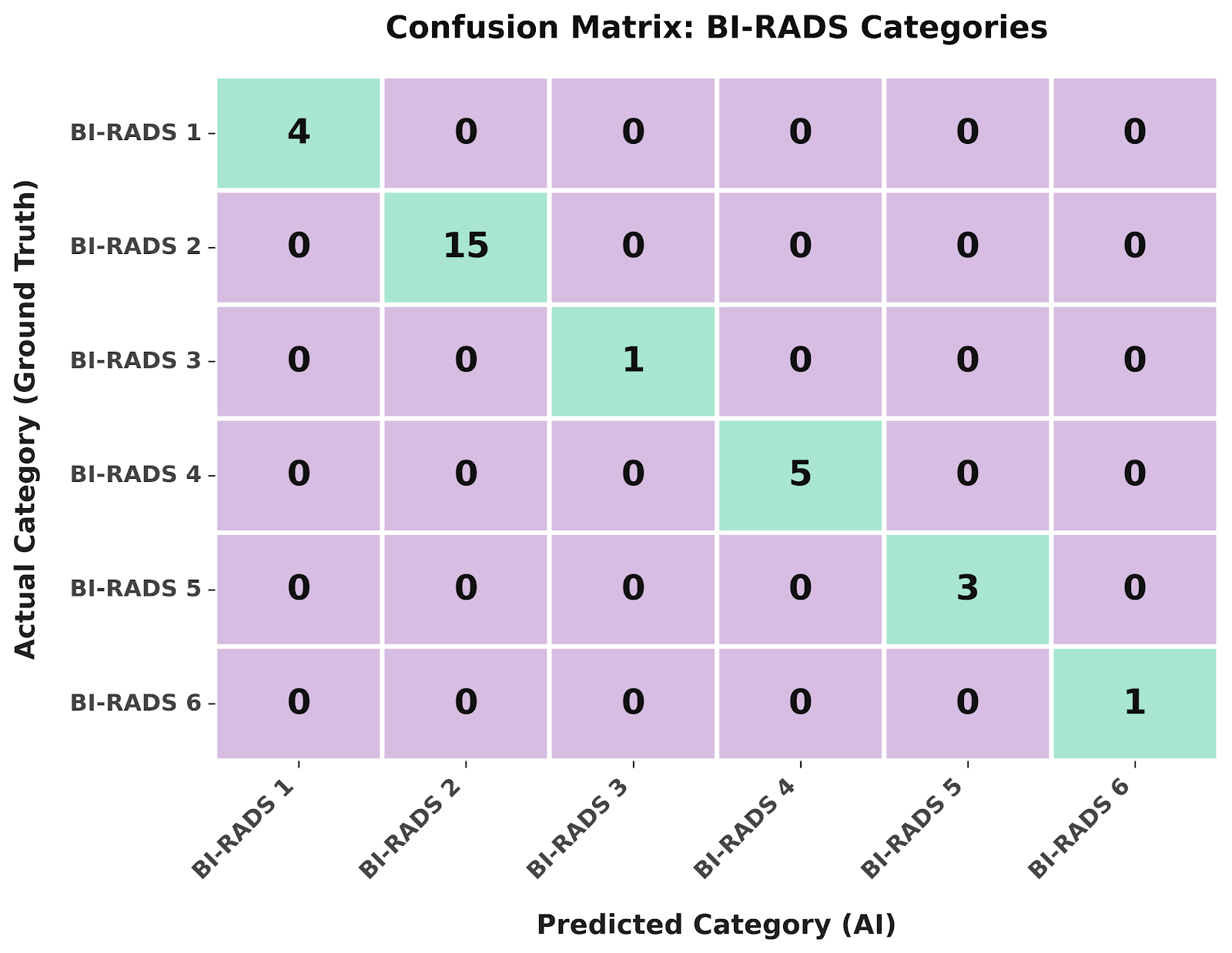}
  \caption{Confusion Matrix for the BI-RADS Category. All
    29~examinations were correctly classified in categories 1 to~6
    (Accuracy\,=\,1.00), with no overlap between classes.}
  \label{fig:confusao_birads}
\end{figure}

As the BI-RADS category constitutes the only multiclass variable (categories
1 to~6), its results are presented separately in
Figures~\ref{fig:confusao_global} and~\ref{fig:confusao_birads}.

Figure~\ref{fig:confusao_global} presents the global confusion
matrix for binary entities. Of the 116~evaluated instances, the model recorded
39~True Positives (TP) and 75~True Negatives (TN), with only 1~False Positive
(FP) and 1~False Negative (FN), evidencing high discriminative capacity for
both the presence and absence of clinical entities.

Figure~\ref{fig:confusao_birads} displays the confusion matrix for
BI-RADS categories 1 to~6. All 29~examinations were correctly classified on the
main diagonal --- BI-RADS~2 was the most frequent category ($n$\,=\,15),
followed by BI-RADS~4 ($n$\,=\,5), BI-RADS~1 ($n$\,=\,4), BI-RADS~5
($n$\,=\,3), and BI-RADS~6 ($n$\,=\,1) ---,
with no confusion between classes, resulting in Accuracy of~1.00.
This result is clinically relevant, as BI-RADS classification errors can
imply inadequate clinical management, from delayed necessary biopsies to
unnecessary interventions.

\FloatBarrier

\FloatBarrier

\subsection{Comparison Between AI and Human}\label{sec:comparacao}

Table~\ref{tab:comparacao} presents the comparative performance between the
human (pre-adjudication manual extraction: TP\,=\,37, TN\,=\,75, FP\,=\,1,
FN\,=\,3) and the Gemini~2.5~Flash model, with the adjudicated ground truth
adopted as the gold standard in both cases.

\begin{table}[htbp]
\centering
\caption{Comparison of F1-Score and Sensitivity per clinical entity
         between human and Gemini~2.5~Flash (adjudicated ground truth as
         gold standard).}
\label{tab:comparacao}
\begin{tabular}{@{} lcccc @{}}
\toprule
\textbf{Clinical Entity}
  & \multicolumn{2}{c}{\textbf{F1-Score}}
  & \multicolumn{2}{c}{\textbf{Sensitivity}} \\
\cmidrule(lr){2-3}\cmidrule(lr){4-5}
  & \textbf{Human} & \textbf{AI}
  & \textbf{Human} & \textbf{AI} \\
\midrule
Cysts                & 1.00 & 1.00 & 1.00 & 1.00 \\
Nodules              & 0.90 & 1.00 & 0.90 & 1.00 \\
Calcifications       & 0.97 & 0.97 & 0.94 & 0.94 \\
Microcalcifications  & 0.00\textsuperscript{a} & 0.66 & 0.00 & 1.00 \\
\midrule
\textbf{Macro F1}    & \textbf{0.72} & \textbf{0.91}
                     & \textbf{0.71} & \textbf{0.98} \\
\textbf{Micro F1}    & \textbf{0.95} & \textbf{0.98}
                     & \textbf{0.93} & \textbf{0.98} \\
\bottomrule
\end{tabular}
\par\smallskip
\footnotesize\textsuperscript{a}Precision undefined (TP\,=\,0, FP\,=\,0);
F1\,=\,0. Human Micro~F1: TP\,=\,37, FP\,=\,1, FN\,=\,3
(Precision\,=\,0.97; Recall\,=\,0.93).
\end{table}

\FloatBarrier

The most expressive difference was concentrated in the Microcalcifications
category. The only real case of this entity in the test set was detected by
the AI (sensitivity\,=\,1.00), while the human did not flag it
(sensitivity\,=\,0.00). The human accumulated three omissions (FN\,=\,3)
against only one by the AI (FN\,=\,1), indicating greater human susceptibility
to omission errors in systematic report review. The AI Macro~F1 (0.91)
outperformed that of the human (0.72) by 0.19~points. The Micro~F1, by
weighting by instance volume, places the performances closer together
(0.98 vs.\ 0.95).

Qualitative analysis of errors and correct extractions revealed that the AI
acted as a verification tool (\textit{double-check}), mitigating human
attention errors. The main observations were:

\begin{itemize}
  \item \textbf{Correction of Human False Positives (Specificity):} In
        Exam~1, manual extraction incorrectly indicated the presence of a
        nodule. The AI correctly identified its absence (specificity\,=\,1.00).

  \item \textbf{Correction of Human Omissions (Sensitivity):} In Exams
        with identifier numbers~10, 16, and~70, the model detected
        calcifications and nodules omitted in the manual tabulation.
        Additionally, the only case of microcalcification in the test set
        was identified by the AI and not by the human.

  \item \textbf{Taxonomic Error:} In Exam~4, the model classified a
        calcification as a microcalcification (dimensional criterion
        $<$ 0.5~cm vs.\ the human morphological criterion), generating
        1~FP and 1~FN. This behavior is correctable through prompt
        adjustment.
\end{itemize}

\subsection{Subjective Validation (Agreement)}\label{sec:concordancia}

\begin{figure}[htbp]
  \centering
  \includegraphics[width=.7\linewidth]{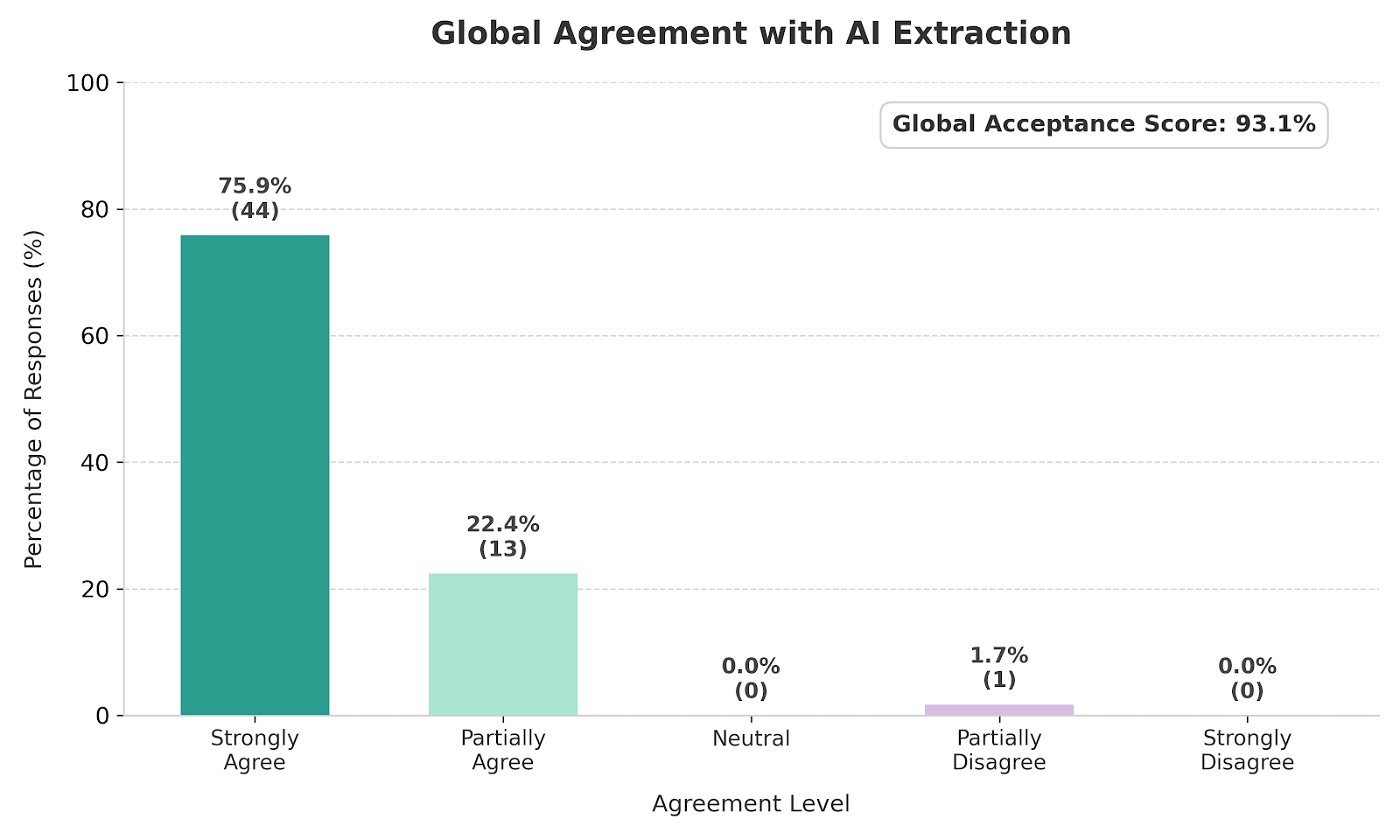}
  \caption{Level of professional agreement with AI. The Global
    Acceptance Index calculated over 58~evaluations (Likert scale, two
    professionals) was 93.1\%.}
  \label{fig:fig3}
\end{figure}

To measure qualitative performance, the Global Acceptance Score was adopted,
converting Likert scale responses to numerical values
(`Strongly Agree'\,=\,100; `Strongly Disagree'\,=\,0). The final index obtained
was 93.1\% (Figure~\ref{fig:fig3}), based on 58 assessments conducted by two evaluators, with each one responsible for 29 exams. 

Of the 58~evaluations, 75.9\% ($n$\,=\,44) corresponded to ``Strongly Agree''
and 22.4\% ($n$\,=\,13) to ``Partially Agree''. Only one evaluation (1.7\%)
indicated partial disagreement, and no total disagreement was recorded.

\FloatBarrier

The results demonstrated that the AI presented significant efficacy in
automating the extraction process, with greater completeness of results, lower
probability of omissions, and superior speed. The professionals demonstrated
advantages in interpretive flexibility for atypical cases, such as Exam~4.

\begin{figure}[htbp]
  \centering
  \includegraphics[width=.8\linewidth]{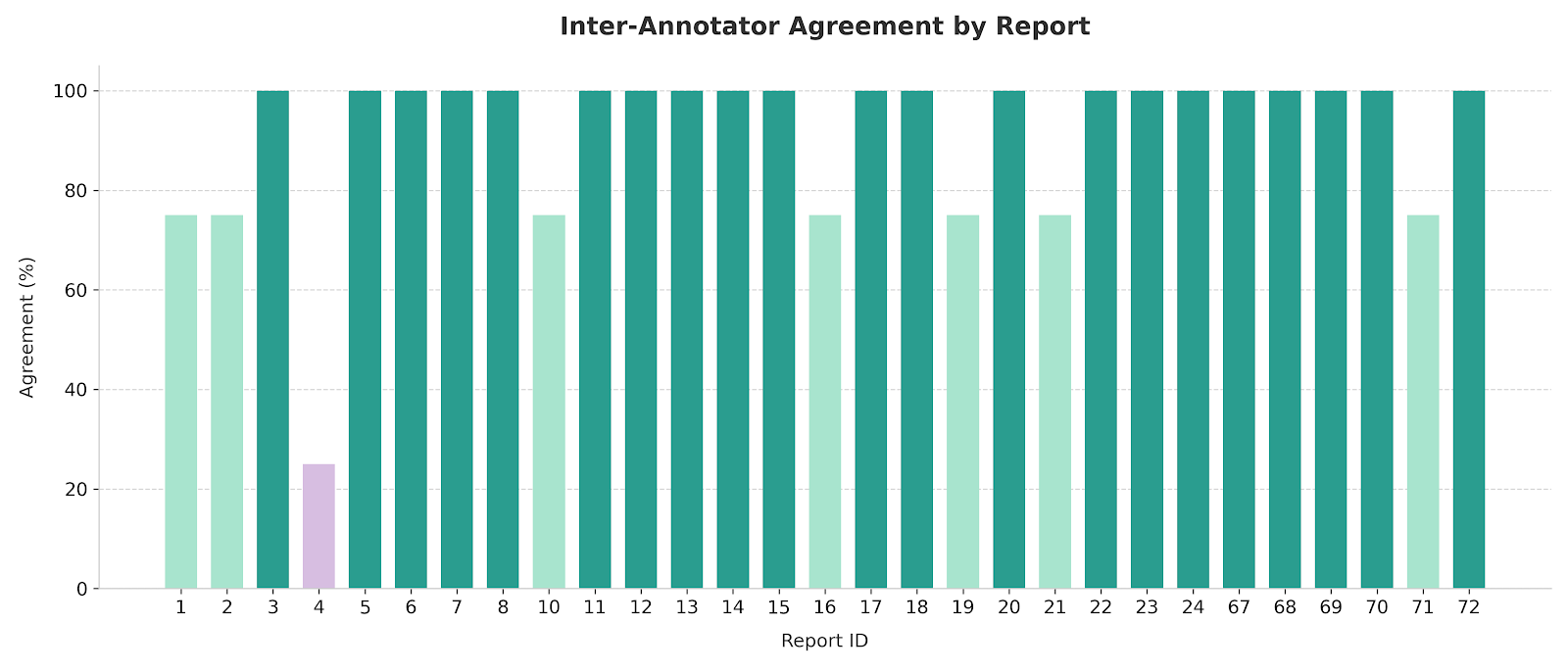}
  \caption{Levels of agreement between professionals by examination.
    Dark green\,=\,full agreement (100\%); light green\,=\,subtle divergence;
    lilac\,=\,disagreement. Only Exam~4 showed substantive divergence regarding
    the punctiform calcification.}
  \label{fig:fig4}
\end{figure}

Figure~\ref{fig:fig4} presents the levels of agreement between professionals
by examination. The color scale reflects the degree of consensus: dark green
indicates full agreement; light green denotes subtle divergence (between
`strongly agree' and `partially agree'); and lilac signals disagreement. Only
in Exam~4 was there interpretive divergence regarding the classification of the
punctiform calcification.

\FloatBarrier

%% ============================================================
\section{Discussion}\label{sec:discussao}
%% ============================================================

\subsection{Error Analysis}\label{sec:erros}

\begin{figure}[htbp]
  \centering
  \includegraphics[width=\linewidth]{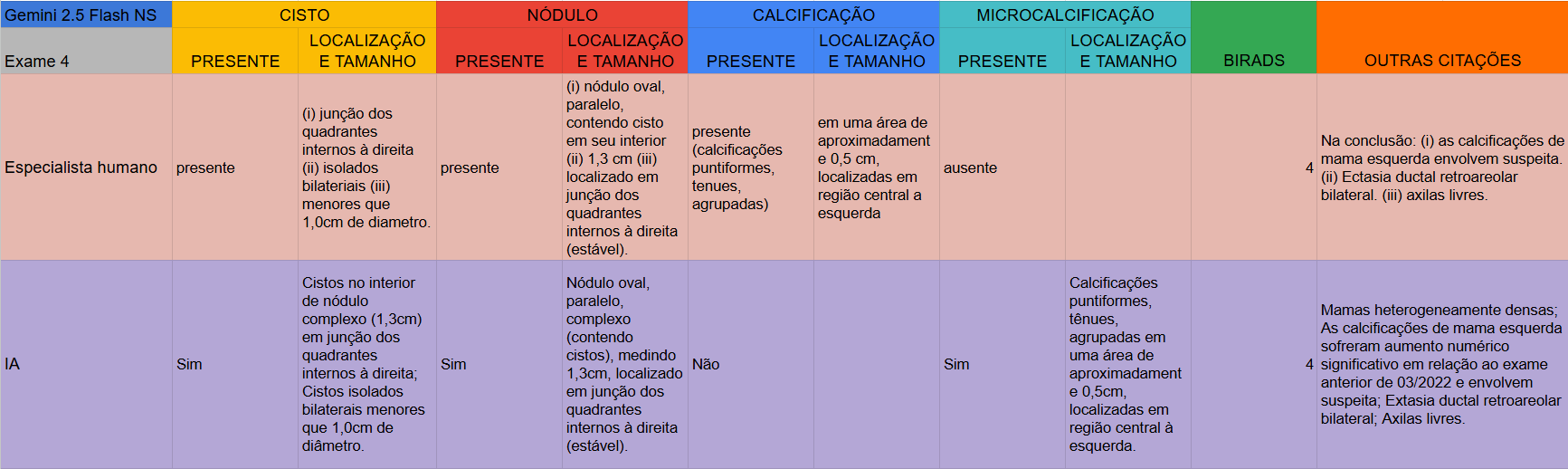}
  \caption{Exam~4: comparison between AI extraction
    (Gemini~2.5~Flash) and human. The AI applied a dimensional criterion
    ($<$ 0.5~cm) to classify the calcification as a microcalcification,
    while the human adopted morphological and clinical criteria. Disagreement
    correctable through prompt adjustment.}
  \label{fig:exame4pt}
\end{figure}

There were two residual errors mitigable through prompt adjustment.
Figure~\ref{fig:exame4pt} details the inconsistency in Exam~4, where
the model classified a punctiform calcification as a microcalcification
(strict size criterion: $<$ 0.5~cm), while the professionals adopted
morphological and clinical criteria. This disagreement is correctable through
prompt adjustments.

\FloatBarrier

\begin{figure}[htbp]
  \centering
  \includegraphics[width=\linewidth]{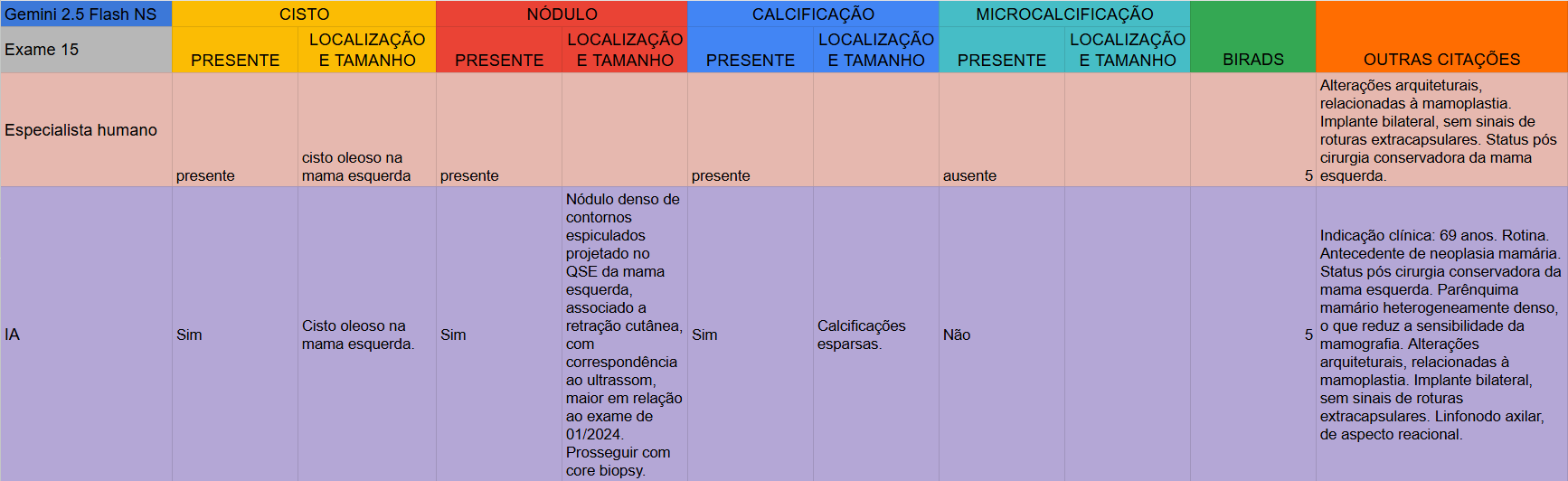}
  \caption{Exam~15: comparison between AI and human extraction.
    Report selected for its tendency to induce hallucinations in PDF format;
    full agreement achieved across all variables with .txt input.}
  \label{fig:exame15pt}
\end{figure}

Exam~15 (Figure~\ref{fig:exame15pt}) was prioritized in tests for
inducing hallucinations in PDF format. In .txt format, full agreement was
achieved across all clinical variables. This result is consistent with the
literature indicating that the quality of input preprocessing directly
influences the fidelity of LLM responses~\cite{ji2023}, and with practical
studies demonstrating that structural elements of PDFs --- such as tables,
metadata, and complex layouts --- represent significant challenges for text
extraction intended for AI pipelines, favoring the use of plain text formats.

\FloatBarrier

\begin{figure}[htbp]
  \centering
  \includegraphics[width=\linewidth]{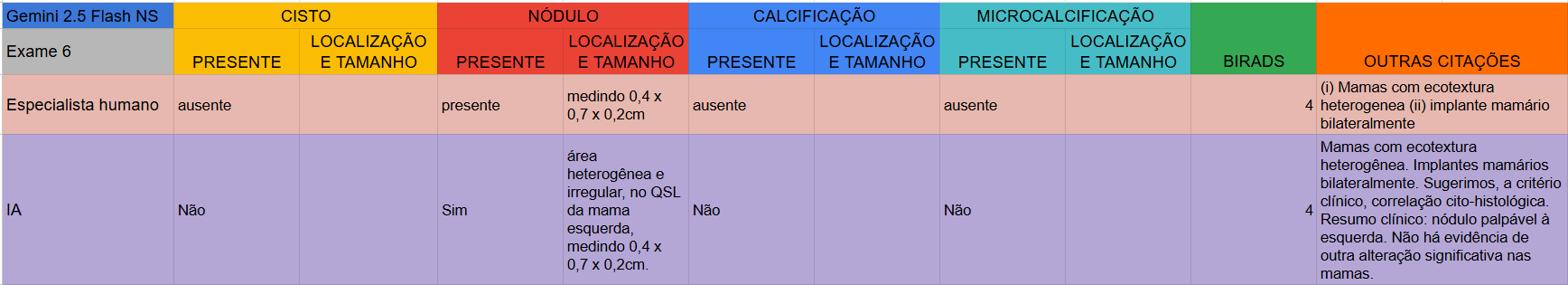}
  \caption{Exam~6: comparison between AI and human extraction.
    Reference report in tests, with full agreement -- absence of cyst,
    calcification, and microcalcifications, and presence of nodule with
    description.}
  \label{fig:exame6pt}
\end{figure}

\begin{figure}[htbp]
  \centering
  \includegraphics[width=\linewidth]{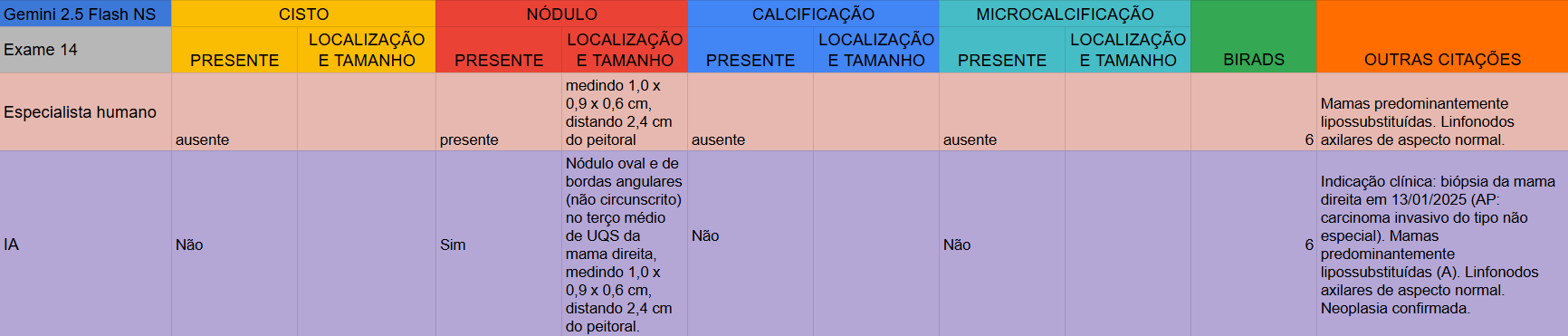}
  \caption{Exam~14: comparison between AI and human extraction,
    demonstrating full agreement including BI-RADS~6 classification and
    complete nodule description.}
  \label{fig:exame14pt}
\end{figure}

Exams~6 and~14 (Figures~\ref{fig:exame6pt}
and~\ref{fig:exame14pt}) are representative of the expected model behavior
under normal conditions. In both cases, the extraction performed by the AI was
fully consistent with the gold standard, correctly identifying the presence and
absence of each clinical entity and the corresponding BI-RADS classification.
These results corroborate the findings of Akcali~\textit{et
al.}~\cite{akcali2025}, who reported Macro~F1 between 0.84 and~0.99 for
similar tasks in mammography reports in non-English languages, indicating that
the proposed approach is competitive even without model fine-tuning and with a
reduced example corpus.

\FloatBarrier

\FloatBarrier

\subsection{Limitations}\label{sec:limitacoes}

The main limitations involved: (1)~a dataset limited to 33~examinations from a
single partner mastology clinic, in compliance with the ethical restrictions
imposed by the General Personal Data Protection Law (LGPD -- Law
No.~13.709/2018); (2)~specificity of the \textit{prompt}, developed and
calibrated from the linguistic and structural patterns of Mammography and Breast
Ultrasound examinations only; and (3)~low adherence of researchers to the
validation form, which was sent to the research team (3 members) and the partner
mastologist, resulting in only two respondents.

%% ============================================================
\section{Conclusion}\label{sec:conclusao}
%% ============================================================

This study concludes that LLMs, when instructed through Prompt Engineering,
demonstrate effective capacity to assist in the extraction and summarization of
clinical findings from mammography and breast ultrasound reports. Prompt
Engineering proved to be a computationally low-cost solution for structuring
semi-structured data in a clinical environment.

The results confirmed the hypothesis that LLMs can achieve performance
comparable to or superior to manual extraction by specialists, as evidenced by
the Macro~F1 of~0.91 and Micro~F1 of~0.98, compared to the human Macro~F1
of~0.72 and Micro~F1 of~0.95. The finding that the AI detected the only case
of microcalcification present in the test set, while the human missed it,
reinforces the model's potential as a complementary \textit{double-check} tool,
especially for low-prevalence entities. However, this omission should not be
interpreted as a limitation exclusive to the professional's technical
competence, but rather as a reflection of the inherent conditions of clinical
practice. Human attention is a finite resource susceptible to cognitive fatigue,
which tends to intensify over long work shifts, in the face of high patient
volumes, or in contexts of institutional pressure. The subjective validation
(Likert scale, 93.1\% acceptance) corroborated these results.

A notable aspect of this study is the model's robustness against real-world
data variability. Since the sample included reports from various laboratories
and radiology services in Rio Grande do Sul, which naturally present different
formatting, nomenclature, and writing styles, the high performance achieved
demonstrates the flexibility of the solution. This suggests that the developed
\textit{prompt} architecture is highly adaptable and has great generalization
potential, being applicable for data extraction from virtually any mammography
and breast ultrasound report written in Brazilian Portuguese, regardless of the
issuing institution.

For subsequent studies, the following are planned: (1)~increasing the data
corpus for more robust validation; (2)~adapting the \textit{prompt} for
different examination types and report standards; and (3)~including a greater
number of health professionals in the validation of AI responses.

\section{Ethics Statement}\label{sec:etica}
This study and the acquisition of reports comply with the General Personal Data
Protection Law (LGPD -- Law No.~13.709/2018). The examinations were obtained
through a Mastology Clinic, with approval by the Research Ethics Committee
(CEP) via the Plataforma Brasil platform, under CAAE
No.~88036825.9.0000.5343. Included were digital reports of mammography and breast
ultrasound examinations that met the following criteria: patient authorization
through signature of the Informed Consent Form (ICF); and compliance with
current data protection regulations.
To preserve patient privacy, personal identification data were removed or
anonymized prior to AI analysis, in accordance with research ethical principles
and the provisions of the LGPD. Reports that were incomplete, in formats
unsuitable for processing, or whose patients did not consent to participation
in the research were excluded.

\section*{Data Availability Statement}
The dataset used in this study consists of mammography and breast ultrasound
reports obtained from a partner mastology clinic under the approval of the Research
Ethics Committee (CEP), CAAE No.~88036825.9.0000.5343, in compliance with the General
Personal Data Protection Law (LGPD -- Law No.~13.709/2018). Due to patient
privacy constraints and applicable data protection regulations, the dataset
is not publicly available. Requests regarding access to the data may be
directed to the corresponding author.

\section*{Declaration of Competing Interests}
The authors declare that Clínica de Mastologia Dra.\ Ana Paula Muller Ltda.\
provided the reports used in this study. Dr.\ Ana Paula Wernz da Cunha Müller served as an
external evaluator in the qualitative validation stage (Likert scale) and
assisted in defining the variable search process together with the health
researchers, without involvement in the remaining stages of the research. The
external collaborators had no influence on the study design, data collection,
analysis and interpretation, manuscript writing, or the decision to submit it
for publication. The remaining authors declare no competing interests.

\section*{Declaration of Generative AI and AI-Assisted Technologies in the Manuscript Preparation Process}
The authors declare that generative AI tools were used in a limited and supervised capacity during the preparation of this manuscript. Specifically, Claude Sonnet 4.6 (Anthropic) was used for grammar correction and linguistic improvement. No portion of the text was generated by AI. All ideas, arguments, and written content are exclusively of human authorship. After each AI-assisted revision, the authors thoroughly reviewed and approved the modified text to ensure its accuracy, coherence, and compliance with the study's objectives.

\section*{Acknowledgements}
\textbf{Funding:} This work was developed within the scope of the project
``Cognitive solution for identifying findings of examinations in the breast
cancer care pathway: screening and pre-diagnosis'', affiliated with the
Universidade de Santa Cruz do Sul (UNISC). The research is financially
supported by the National Council for Scientific and Technological Development
(CNPq) and the Department of Science and Technology (Decit/SECTICS/MS),
under CNPq/Decit Call No.~21/2023 -- Transdisciplinary Studies in
Collective Health.

%% ============================================================

\section*{CRediT Authorship Contribution Statement}
\textbf{Lorenzo Farias:} Conceptualization, Methodology, Software, Validation, Writing -- Original Draft.
\textbf{Hanna Reckziegel:} Conceptualization, Methodology, Software, Validation, Writing -- Original Draft.
\textbf{Daniela Duarte da Silva Bagatini:} Conceptualization, Methodology, Supervision, Validation, Writing -- Review \& Editing.
\textbf{Daniel Schulz:} Validation.
\textbf{Gabriela de Andrade Monteiro:} Methodology, Writing -- Review \& Editing.
\textbf{Letícia Zanatta:} Validation.
\textbf{Ana Laura Brill Thum:} Validation.
\textbf{Priscila Schmidt Lora:} Methodology, Validation.
\textbf{Débora Oliveira da Silva:} Supervision, Funding Acquisition.
\textbf{Ana Paula Wernz da Cunha Müller:} Validation, Resources.
\textbf{Cristiane Drebes Pedron:} Validation.

%% ============================================================
\bibliographystyle{unsrtnat}
\bibliography{references}

\end{document}